\documentclass{article}
\usepackage{spconf}
\usepackage{amssymb,amsfonts}
\usepackage{amsmath,graphicx}
\usepackage[ruled,vlined]{algorithm2e}
\usepackage{caption}
\usepackage[]{hyperref}
\usepackage{multirow}
\usepackage{float}
\usepackage{booktabs}
\def\x{{\mathbf x}}

\newcommand{\ie}{\textit{i.e.}\xspace}
\newcommand{\eg}{\textit{e.g.}\xspace}
\newcommand{\xhdr}[1]{\vspace{1.3mm}\noindent{{\bf #1.}}}

\title{Node Attribute Completion in Knowledge Graphs with Multi-relational Propagation}
\name{Eda Bayram* \thanks{*Corresponding author: eda.bayram@epfl.ch}\qquad Alberto Garc\'{i}a-Dur\'{a}n \qquad Robert West} 
\address{EPFL}
\begin{document}
\ninept
\maketitle
\begin{abstract}
The existing literature on knowledge graph completion mostly focuses on the link prediction task. However, knowledge graphs have an additional incompleteness problem: their nodes possess numerical attributes, whose values are often missing.
Our approach, denoted as \textsc{MrAP}, imputes the values of missing attributes by propagating information across the multi-relational structure of a knowledge graph.
It employs regression functions for predicting one node attribute from another depending on the relationship between the nodes and the type of the attributes.
The propagation mechanism operates iteratively in a message passing scheme that collects the predictions at every iteration and updates the value of the node attributes. Experiments over two benchmark datasets show the effectiveness of our approach.
\end{abstract}
\section{Introduction}
\label{sec:intro}
Knowledge graphs (KGs) consist of structured data formed by semantic entities connected through multiple types of relationships. They play an important role in a wide variety of AI applications including question answering \cite{west2014knowledge, bordes2014question}, drug discovery \cite{mohamed2020discovering, ioannidis2020few}, and e-commerce \cite{xu2020product, li2020alime}. In the last years, this has led to immense attention on knowledge graph completion methods, which aim at inferring missing facts in a KG by reasoning about the observed facts \cite{nickel2015review}. Knowledge graph embedding (KGE) methods are at the core of this progress, by learning latent representations for both entities and relations in a KG \cite{wang2017knowledge}.
In relational representation learning, graph neural network (GNN) \cite{scarselli2008graph} and message passing neural network (MPNN) \cite{gilmer2017neural} methods have also been effectively used. While originally these methods were designed for simple undirected graphs, there are also works that incorporate multi-relational information \cite{schlichtkrull2018modeling}.

\begin{figure}[t]
  \centering
  \captionsetup{justification=centering}
  \includegraphics[width=1\columnwidth]{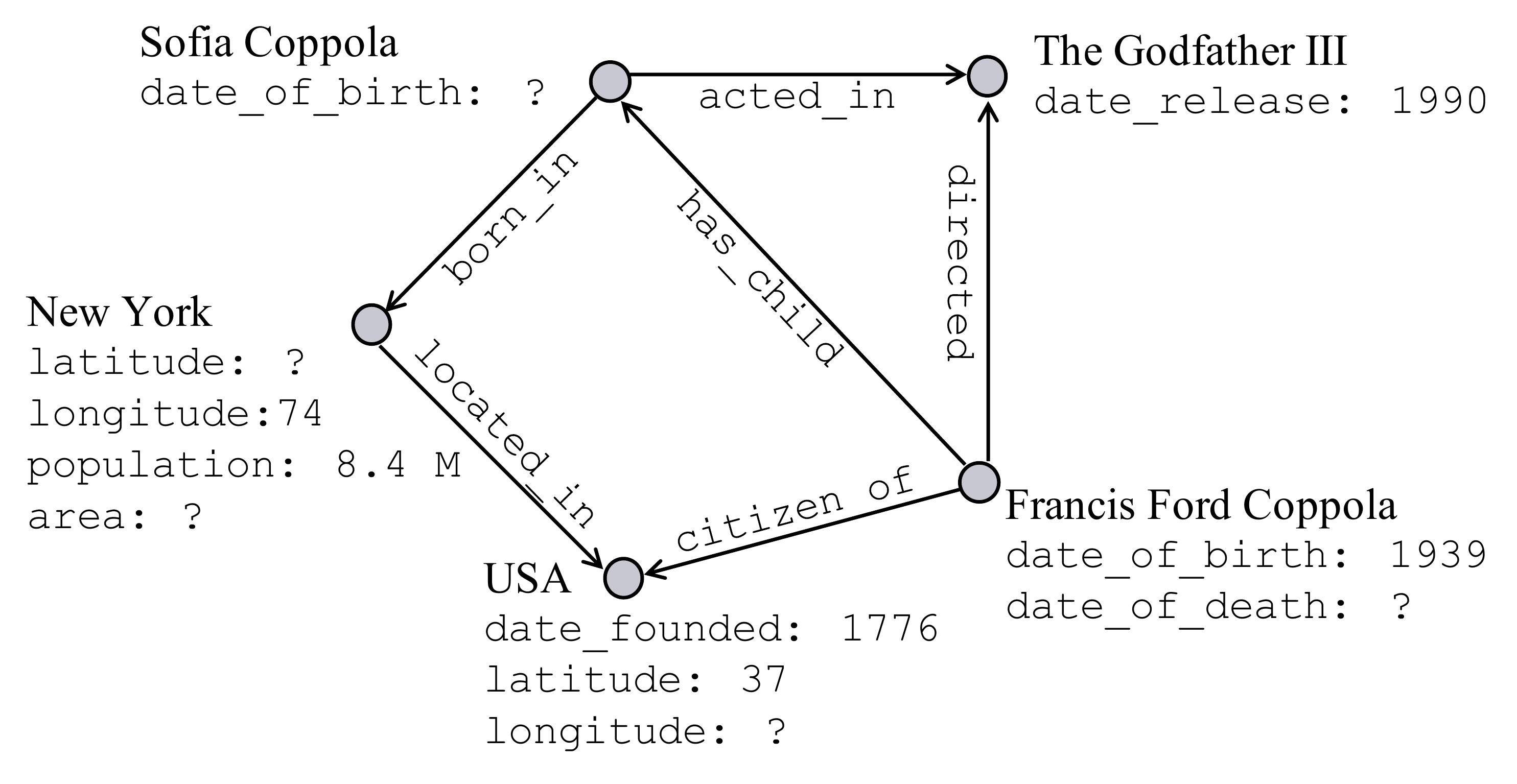}
	\caption{A part of KG data with incomplete node attributes}
  \label{fig:missingKB}
\end{figure}

Despite the very large number of KGE and GNN methods, these works have mostly addressed link prediction and node/graph classification problems, respectively. While KGE methods always harness features learned from the relational structure of the graph, they very often overlook other information contained in the KGs such as the numerical properties of the entities. In this work, we shift the focus away from the aforementioned problems, and study the much less explored problem of node attribute prediction in KGs.
Here, we particularly address the incompleteness in the numerical node attributes that are expressed in continuous values.
Figure \ref{fig:missingKB} depicts an example: the node New York does not have a value for two numerical attributes, \texttt{latitude} and \texttt{area}, it should possess. Similarly, we observe missing values in some attributes of other nodes of the KG. Node attribute completion is the task of finding appropriate values for the nodes' numerical attributes that do not have an annotated value.

Different to the standard KG completion problems, in node attribute completion task, we harness not only the relational structure
of the graph, but also the correlation between various types of node attributes.
First of all, the relational structure provides very rich predictive information. As seen in Figure \ref{fig:missingKB}, one may provide estimates for some missing attributes given the known ones at the neighboring nodes and the relationships that hold between them.
Therefore, in this study, we impute the values of missing attributes by propagating information across the multi-relational structure of the KG. Second, the prediction of missing node attributes also depends on their correlation to the attribute types observed at the neighboring nodes.
Thus, we employ a number of regression functions that predict an attribute of a node from an attribute of its neighbor with respect to both the type of the attributes and the relation between the nodes. We also adopt another set of regression functions for the pair of attributes that can be accommodated at the same node, \eg , for predicting \texttt{date}$\_$\texttt{of}$\_$\texttt{death} from \texttt{date}$\_$\texttt{of}$\_$\texttt{birth} within a node. In addition, we assign a weight to each regression function reflecting its predictive power, which will be taken into account during the propagation of their predictions. The parameters of the regression functions and the weights are estimated based on the observed set of node attributes prior to the propagation procedure.

\xhdr{Related Work}
Although many KGs often contain numerical properties attributed to entities, very few studies have explored and exploited them \cite{garcia2017kblrn, kotnis2018learning, shang2019end}.
The numerical attribute prediction problem was recently introduced by Kotnis and Garcia-Duran \cite{kotnis2018learning}, who address the problem with a two-step framework called \textsc{Nap++}. First, they extend the KGE approach to learn node embeddings underlying a KG enriched with numerical node attributes. Second, they build a k-NN graph upon the embedding to propagate the known values of node attributes towards the missing ones. Propagating information on a surrogate graph constructed on the embedding is rather sub-optimal compared to leveraging the original relational structure of the KG. As opposed to that, in this study, we propose a propagation algorithm
that directly operates on the inherent structure of the KG. For this purpose, we take inspiration from the well-known label propagation algorithm \cite{zhu2002learning}, which infers the label of a node from its neighbors iteratively under the assumption that nearby nodes should have similar values. However, this technique is insufficient to handle the complexity of KGs, which possess multiple types of attributes and multiple types of relationships following different affinity rules between neighboring nodes. For example, two nodes linked via the relationship \texttt{has}$\_$\texttt{child} exhibit a certain bias between their \texttt{date}$\_$\texttt{of}$\_$\texttt{birth} attributes, but do not necessarily have similar values.
The authors in \cite{garcia2017kblrn} exploit such numerical node attributes in a KG for the multi-relational link prediction task. Instead of adopting the plain difference between the values of neighboring node attributes, they model the affinity using a radial basis function, which can account for the aforementioned bias term that may arise in some relations.
Similarly in our method, the introduced regression functions model a linear relation between neighboring node attributes.
Moreover, the regression functions are able to model the linear correlation between different types of attributes.
Therefore, our method allows propagation between node attributes of different types, unlike the previous numerical attribute propagation solution \cite{kotnis2018learning}. The GNN and MPNN methods also learn node representations by propagating them along the edges of a graph. Recently, multi-relational variants have also been developed, which usually augment the learning parameters in a relation-specific manner \cite{li2015gated, hamaguchi2017knowledge, schlichtkrull2018modeling, beck2018graph, brockschmidt2019gnn, teru2019inductive, vashishth2019composition, yu2020generalized}. The main difference of the proposed method from those is that it propagates incomplete node features across the graph instead of propagating fixed dimension of node representation vectors.
Another line of work exploiting the multi-relational structure of a graph learns mask coefficients or attention weights for multiple types of edges \cite{neil2018interpretable, wang2019kgat, busbridge2019relational, shang2019end, bayram2020mask, hu2020heterogeneous}. This approach enables discriminating the importance of the neighboring nodes for the inference task, rather than treating them equally.
Similar to an attention mechanism, in our method, the assigned weights for the regression functions capture the importance of the collected predictions for a certain attribute in a node.

\xhdr{Contributions}
In this study, we propose a multi-relational attribute propagation algorithm, \textsc{MrAP}, which directly operates on the original structure of the knowledge graph. \textsc{MrAP} imputes missing numerical attributes by iteratively applying two steps to each node attribute: it collects all predictions about the node attribute and updates its value by aggregating the predictions based on their weights. We formulate \textsc{MrAP} within a message passing scheme described in \cite{xu2018representation}. To the best of our knowledge, we are the first one to realize message passing with incomplete heterogeneous node features and demonstrate its applicability for the node attribute completion task. Experiments show its superior performance in two KG datasets.

\section{Multi-Relational Attribute Propagation Algorithm}
\begin{figure}[t]
  \centering
  \captionsetup{justification=centering}
   \includegraphics[width=1\columnwidth]{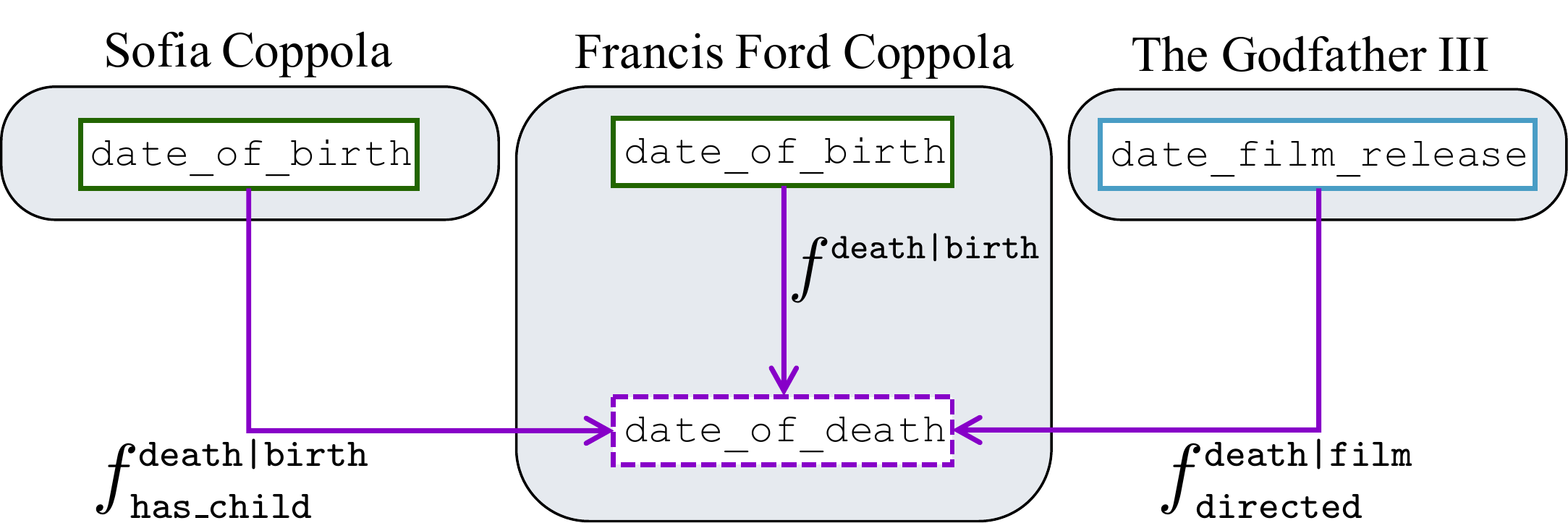}
	\caption{Message passing performed by \textsc{MrAP} to update the attribute \texttt{date\_of\_death} for the node Francis Ford Coppola.}
  \label{fig:message_pass}
\end{figure}
\xhdr{Notation} A KG enriched with node attributes is denoted as $\mathcal{G} = (\mathcal{V}, \mathcal{E}, \mathcal{P}, \mathcal{A})$, where $\mathcal{V}$ is the set of nodes (entities), $\mathcal{P}$ is the set of relation types, $\mathcal{E} \subseteq  \mathcal{V} \times  \mathcal{P} \times \mathcal{V}$ is the set of multi-relational edges, and $\mathcal{A}$ is the set of attribute types. Moreover, $\mathcal{N}_v$ is the set of all neighbors of node $v \in \mathcal{V}$, and $\mathcal{A}_v$ is the set of attributes belonging to $v$.
The function $\mathtt{r}(v,n)$ returns the relation type $\mathtt{p} \in \mathcal{P}$ that is pointed from node $n$ to node $v$. If such a relation exists between them, yet pointed from the node $v$ to the node $n$, then the function returns the reverse as $\mathtt{p}^{-1}$. In addition, we denote $x_n$ for the value of attribute $x$ belonging to node $n$, \ie, $x \in \mathcal{A}_n$.

\xhdr{Approach} \textsc{MrAP} explicitly makes use of the multi-relational structure given by the KG and the observed numerical node attributes to infer the missing ones. Humans also use these inputs to perform numerical reasoning. For instance, in Figure \ref{fig:missingKB}, one may provide an estimate about the date of death of Francis Ford Coppola by looking at the release date of one of his most popular movies. For this purpose, we introduce a number of regression functions denoted as $f^{y | x}_\mathtt{p}$ to predict an attribute of type $y$ from an attribute of type $x$ through a relation of type $\mathtt{p}$ that holds between the nodes accommodating $y$ and $x$ respectively. The date of death of Francis Ford Coppola can also be estimated from his own date of birth. Accordingly, we employ another set of regression functions denoted as $f^{y | x}$ in order to predict attribute $y$ from attribute $x$ within a node accommodating both of the attributes at the same time. $f^{y | x}$ is obviously relation independent.
In addition, humans have the capacity to determine the predictive power of each source of information, and weight each information accordingly in their numerical reasoning process. Similarly in our approach, each regression function is assigned with a weight denoted by $\omega^{y | x}_\mathtt{p}$ (or $\omega^{y | x}$), reflecting its predictive power. The proposed method, \textsc{MrAP}, recovers the values of missing node attributes by minimizing their distances to the predictions collected from such internal and external sources of information based on their weights. For an arbitrary node $v \in \mathcal{V}$, we define this loss as
\begin{equation}
\begin{split}
\mathcal{L}_v = \sum_{y \in \mathcal{A}_v} \bigg(&\underbrace{\sum_{n \in \mathcal{N}_v} \sum_{x \in  \mathcal{A}_n} w^{y | x}_{\mathtt{r}(v,n)} d(y_v, f^{y | x}_{\mathtt{r}(v,n)}(x_n))}_{\text{outer loss}} \\
     & \quad + \underbrace{\sum_{\substack{x \in  \mathcal{A}_v\\
     x \neq y}} w^{y | x} d(y_v, f^{y | x}(x_v)}_{\text{inner loss}}) \bigg)
    \end{split}
\label{eq:loss}
\end{equation}
where the outer loss accounts for the predictions about $y_v$---attribute $y$ of node $v$---computed with the attributes from the neighboring nodes. On the other hand, the inner loss accounts for the predictions about $y_v$ computed with the attributes within the same node.
The distance function $d$ is computed between $y_v$ and a prediction yielded by the regression functions. The regression functions are specific to the dependent and independent attribute types, $y$ and $x$ respectively. In the outer loss term, the function $f^{y|x}_\mathtt{\mathtt{r}(v,n)}: \mathbb{R} \to \mathbb{R}$ is applied to an explanatory variable $x_n$ where the independent attribute $x$ appears at a neighboring node $n$ connected by the relation $\mathtt{r}(v,n)$. In the inner loss term, on the other hand, the function $f^{y|x}: \mathbb{R} \to \mathbb{R}$ is applied to another attribute $x$ than the dependent attribute type, $y$, encountered at the same node $v$.
The distances are multiplied by the corresponding weights of the regression functions, which controls their contribution in the loss. The loss in \eqref{eq:loss} leads to a least squares problem if the distance function $d$ is simply chosen as squared difference. Then, its solution can be easily derived from $y_v \leftarrow \partial \mathcal{L}_v / \partial y_v = 0$, which is found to be weighted and normalized sum of the predictions yielded by the regression functions.

Ultimately, our learning objective is formalized as the minimization of the loss in \eqref{eq:loss} at every node of the graph \ie,  $ \sum_{v \in \mathcal{V}} \mathcal{L}_v$. Therefore, we design our propagation algorithm \textsc{MrAP} proceeding in two steps that are repeated for a certain number of iterations or until a convergence threshold is reached. First, for each node $v$ and each of its numerical attribute $y$, the node aggregates all messages that aim at predicting $y_v$. Figure \ref{fig:message_pass} illustrates the messages collected by the node Francis Ford Coppola that predicts his $\texttt{date\_of\_death}$ attribute. In a second step, \textsc{MrAP} updates the value of $y_v$ using the collected messages. Based on the solution of \eqref{eq:loss}, the contribution of each prediction for the update of $y_v$ is controlled by its corresponding weight. Then at $k$-th iteration, the new estimate $\hat{y}_v$ is given by:
\begin{equation}
\begin{split}
\hat{y}_v =  \bigg( \sum_{n \in \mathcal{N}_v} \sum_{x \in  \mathcal{A}_n} w^{y | x}_{\mathtt{r}(v,n)}f^{y | x}_{\mathtt{r}(v,n)}(x_n^{k-1}) + \\ \sum_{\substack{x \in  \mathcal{A}_v\\
     x \neq y}} w^{y | x} f^{y | x}(x_v^{k-1}) \bigg) / Q
\end{split}
\label{eq:aggr}
\end{equation}
where $Q =\sum_{n \in \mathcal{N}_v} \sum_{x \in  \mathcal{A}_n} w^{y | x}_{\mathtt{r}(v,n)}+\sum_{\substack{x \in  \mathcal{A}_v\\
     x \neq y}} w^{y | x}$  is a normalization factor, \ie, sum of the weights of the collected predictions. The new estimate is combined with the previous value $y^{k-1}_v$ of the node attribute via a damping factor $\xi$ as follows
     \begin{equation}
         y^{k}_v = (1- \xi) y^{k-1}_v + \xi \hat{y}_v.
    \label{eq:comb}
     \end{equation}
At each iteration, while the values of all missing attributes are updated, the values of a priori known attributes are clamped.

While \textsc{MrAP} imputes the missing node attributes by iteratively applying Eq. \eqref{eq:aggr} and \eqref{eq:comb}, the regression functions and their associated weights are computed in advanced, and kept fixed during the propagation process.

\xhdr{Regression Functions} Each function $f^{y | x}_{\mathtt{r}(v,n)}$ (and $f^{y | x}$) is chosen to be a linear regression function, although more complex functions are also possible. Thus, the regression functions model a linear relationship between the dependent and independent attribute as follows:
\begin{equation}
\label{eqn: regressionModel}
y_v = \eta^{y | x}_{\mathtt{r}(v,n)} x_n + \tau^{y | x}_{\mathtt{r}(v,n)} + \epsilon,
\end{equation}
where $\epsilon \sim N(0, (\sigma^{y | x}_{\mathtt{r}(v,n)})^2)$, \ie the error is normally distributed with a standard deviation of $\sigma^{y | x}_{\mathtt{r}(v,n)}$.
We empirically observed that such linear dependency holds very often between the attributes found in knowledge bases such as DBpedia or Freebase. For instance, the attribute \texttt{date\_of\_birth} of a node can be estimated through a certain value difference from that of a neighbor connected via the relation type \texttt{has\_child}. This motivates the usage of the bias parameter $\tau$. On the other hand, the attributes can be expressed in different units or ranges, for instance, \texttt{weight} of a node can be guessed with a linear correlation to its \texttt{height}, which motivates the parameter $\eta$.
Accordingly, the functions predicting one node attribute from another follow a simple linear regression model, \ie, linear regression with single explanatory variable:
\begin{align}
\label{eqn: regressionFnc}
f^{y | x}_{\mathtt{r}(v,n)}(x_n) =& \: \eta^{y | x}_{\mathtt{r}(v,n)} x_n + \tau^{y | x}_{\mathtt{r}(v,n)},\\
\label{eqn: regressionFnc_inner}
f^{y | x}(x_v) =& \: \eta^{y | x} x_v + \tau^{y | x}.
\end{align}

Note that the multi-relational GNN works mentioned in Section \ref{sec:intro} usually apply a relation specific transformation to the embedding of a node to regress a feature of a neighboring node. The embedding vector is typically composed of all node features. In our case, however, we do not have a fixed dimension of node feature vector, where the number of attributes assigned to each node varies. Thus, we choose to regress one existing node attribute from another in a pairwise manner.

As seen in Eq. \eqref{eq:aggr}, the estimate is obtained by multiplying the predictions with the weight parameter of the corresponding regression function. The variance of the error in \eqref{eqn: regressionModel} relates to the uncertainty of the regression model. For this reason, we directly set the weight parameter as the inverse of the error variance:
$w^{y | x}_{\mathtt{r}(v,n)} = 1/(\sigma^{y | x}_{\mathtt{r}(v,n)})^2$. Therefore, predictions are weighted with respect to the expected error of the corresponding regression function. 

\xhdr{Estimation of model parameters}
It is possible to derive the best fitting values for the parameters of a simple linear regression model from the samples of the dependent and independent variables \cite{rencher2012}. Thus, the parameters of the regression functions are estimated from the observed set of node attributes.
Let $\mathcal{E}_{\mathtt{p}}^{(y,x)}$ be the set of pairs of nodes $(v, n)$ where the relation type $\mathtt{p}$ is pointed from node $n$ to node $v$, and for which the attributes $y$ and $x$ are observed in nodes $v$ and $n$, respectively. We estimate the parameters of the regression function $f^{y | x}_{\mathtt{p}}$ as follows:
\begin{equation}
\label{eqn:eta}
\eta^{y | x}_{\mathtt{p}} = \frac{\sum\limits_{(v,n) \in \mathcal{E}_{\mathtt{p}}^{(y,x)}} (y_v- \mu^y)(x_n- \mu^x)}
{\sum\limits_{(v,n) \in \mathcal{E}_{\mathtt{p}}^{(y,x)}}(x_n- \mu^x)^2},
\end{equation}
where $\mu^x$ is the mean of attribute $x$. Consequently,
\begin{equation}
\label{eqn:tau}
\tau^{y | x}_{\mathtt{p}} = \mathrm{mean}(\{(y_v - \eta^{y | x}_{\mathtt{p}} \x_n) \: | (v,n) \in \mathcal{E}_{\mathtt{p}}^{(y,x)}\}),
\end{equation}
\begin{equation}
\label{eqn:sigma}
(\sigma^{y | x}_{\mathtt{p}})^2 = \mathrm{mean}(\{(y_v - \eta^{y | x}_{\mathtt{p}} \x_n - \tau^{y | x}_{\mathtt{p}})^2 \: | (v,n) \in \mathcal{E}_{\mathtt{p}}^{(y,x)}\}).
\end{equation}

Now, suppose that over the same set of node pairs, $\mathcal{E}_{\mathtt{p}}^{(y,x)}$, we would like to predict $x$ from $y$ with the inverse relationship $\mathtt{r}(n,v) = \mathtt{p}^{-1}$. Then, we rewrite the linear model by reversing the relation in \eqref{eqn: regressionModel}:
\begin{equation}
\label{eqn: regressionRev}
x_n = \cfrac{1}{\eta^{y | x}_{\mathtt{r}(v,n)}}y_v  - \cfrac{\tau^{y | x}_{\mathtt{r}(v,n)}}{\eta^{y | x}_{\mathtt{r}(v,n)}}  - \cfrac{1}{\eta^{y | x}_{\mathtt{r}(v,n)}}\epsilon,
\end{equation}
where the model parameters are diverted and the standard deviation of the error is rescaled by the factor of $\eta^{y | x}_{\mathtt{r}(v,n)}$.
Accordingly, the parameters of function $f^{x | y}_{\mathtt{p}^{-1}}$ regressing $x$ from $y$ through the reverse direction will correspond to:

\begin{equation}
\label{eqn: reverse}
\eta^{x | y}_{\mathtt{p}^{-1}} = \cfrac{1}{\eta^{y | x}_{\mathtt{p}}}, \quad
\tau^{x | y}_{\mathtt{p}^{-1}} = \cfrac{-\tau^{y | x}_{\mathtt{p}}}{\eta^{y | x}_{\mathtt{p}}}, \quad
w^{x | y}_{\mathtt{p}^{-1}} = \cfrac{(\eta^{y | x}_{\mathtt{p}})^2}{(\sigma^{y | x}_{\mathtt{p}})^2}.
\end{equation}

Next, the parameters of the regression functions of the inner loss, $f^{y|x}$, are computed by following a similar procedure. Let $\mathcal{V}^{(y,x)}$ denote the set of nodes for which both the attributes $y$ and $x$ are observed as $y_v$ and $x_v$ respectively. In Eq. \eqref{eqn:eta} and \eqref{eqn:tau}, we replace $\mathcal{E}_{\mathtt{p}}^{(y,x)}$ by $\mathcal{V}^{(y,x)}$ in order to estimate the parameters of the regression function given in \eqref{eqn: regressionFnc_inner}.
Then, the parameters of the regression function $f^{x|y}$, predicting $x$ from $y$, can also be computed using the relations in \eqref{eqn: reverse}.

We finally note that if the linear dependency described in \eqref{eqn: regressionModel} does not exist between a pair of attributes, it is possible to exclude it from \textsc{MrAP}. For this purpose, upon estimating the model parameters, one can check whether the normal error assumption is fulfilled or not.

\xhdr{\textsc{MrAP} as an instance of the MPNN framework}
\textsc{MrAP} can be framed within the forward pass of a Message Passing Neural Network (MPNN) \cite{gilmer2017neural}. The framework defines two generic functions that are used by most of (if not all) the GNNs in the literature. The function ${\scriptstyle \text{AGGREGATE}}$ collects all messages sent to a node and aggregates them. The function ${\scriptstyle \text{COMBINE}}$ takes the aggregated representation and the previous state of the representation to output a new state. In our approach, functions ${\scriptstyle \text{AGGREGATE}}$ and ${\scriptstyle \text{COMBINE}}$ correspond to Eq. (\ref{eq:aggr}) and (\ref{eq:comb}), respectively. The workflow of \textsc{MrAP} is described in Algorithm \ref{alg} using MPNN terminology.

\begin{algorithm}
 \textbf{Input:} $\mathcal{G} = (\mathcal{V}, \mathcal{E}, \mathcal{P}, \mathcal{A})$, regression functions with their associated weights\\
 \textbf{Output:} Imputed node attributes\\
 \textbf{Initialization:} $x^0_n = x_n$ for a priori known attributes \\
 \For{$\text{Until Convergence}$}{
 \For{$y \in \mathcal{A}_v, \forall v \in \mathcal{V} $}{
 $\hat{y}_v =
 {\scriptstyle \text{AGGREGATE}}(
\{ x_n^{k-1} | {\substack{n \in \mathcal{N}_v, \\ x \in  \mathcal{A}_n}}\}
\cup \{ x_v^{k-1} | \substack{ x \in  \mathcal{A}_v \\ x\neq y }\}
)$ \\
 $y_v^k = {\scriptstyle \text{COMBINE}}\big( y_v^{k-1}, \hat{y}_v \big)$
 }
 \text{Clamp a priori known node attributes}}
 \caption{\textsc{MrAP}}
 \label{alg}
\end{algorithm}
\section{Experiments}
\label{sec: Experiments}
\begin{figure*}[t]
  \centering
  \captionsetup{justification=centering}
   \includegraphics[width=1.5\columnwidth]{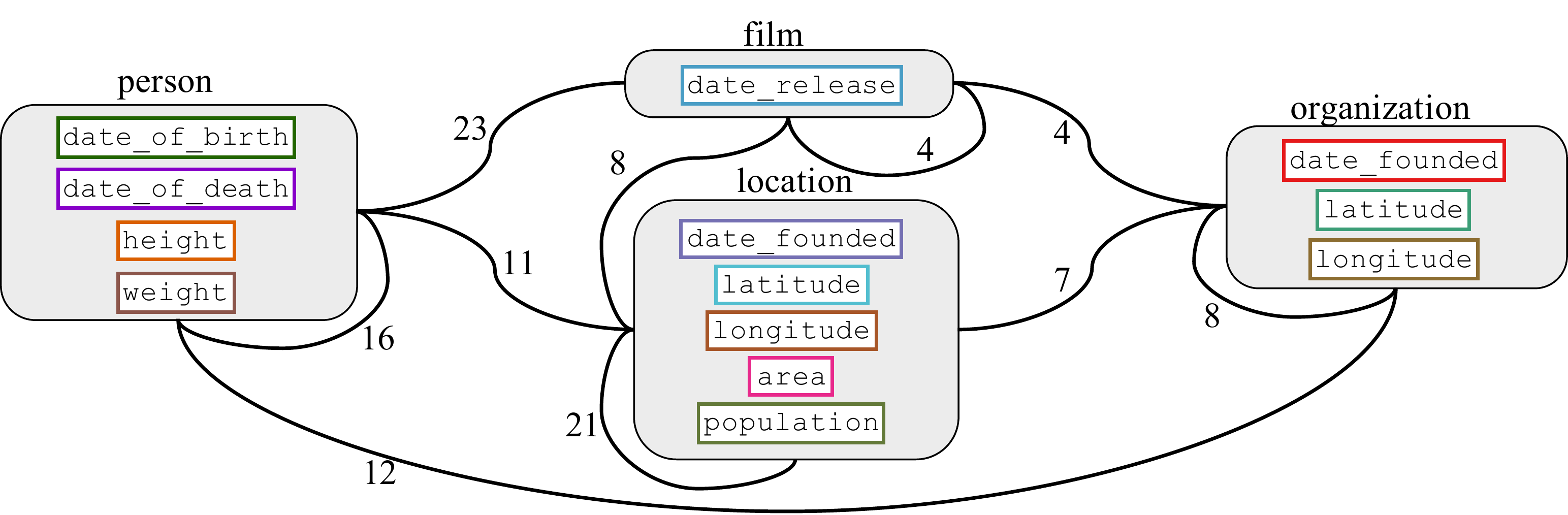}
	\caption{ A summary of FB15K-237 with entity types and numerical attributes encountered on them. The number attached to the connection between a pair of entity types indicates the number of relationship types between those entities.}
  \label{fig: fb_atts}
\end{figure*}
We evaluate the performance of the proposed method on two KG datasets whose nodes have numerical attributes: FB15K-237 \cite{toutanova2015observed} and YAGO15K \cite{garcia2018learning}.
In order to illustrate the complexity of the data, we summarize FB15K-237 dataset in a diagram given in Figure \ref{fig: fb_atts} with the attribute types of interest in the experimental study and the types of entities accommodating those. The number of node attributes of each type encountered in each dataset are also listed in Table \ref{tbl: num_atts}.
Two error metrics are used to assess the performance: Mean Absolute Error (MAE) and Root Mean Square Error (RMSE), which are measured on each type of attribute individually.
\begin{table}[h]
\centering
\caption{\label{tbl: num_atts} Number of node attributes encountered in datasets for each attribute type. The upper block contains numerical attributes of date type. The lower block contains all other attributes. A dash (-) indicates the corresponding attribute is not encountered in the dataset.}
\resizebox{0.7\columnwidth}{!}{
\begin{tabular}{p{3cm}rr}
\toprule
 Attribute                     & FB15K-237 & YAGO15K \\
\midrule                      
\texttt{date\_of\_birth}       & 4406      & 8217    \\
\texttt{date\_of\_death}       & 1214      & 1821    \\
\texttt{film\_release}         & 1853      & -       \\
\texttt{organization\_founded} & 1228      & -       \\
\texttt{location\_founded}     & 917       & -       \\
\texttt{date\_created}         & -         & 6574    \\
\texttt{date\_destroyed}       & -         & 536     \\
\texttt{date\_happened}        & -         & 388     \\
\midrule
\texttt{latitude}              & 3190      & 2989    \\
\texttt{longitude}             & 3192      & 2989    \\
\texttt{area}                  & 2154      & -       \\
\texttt{population}            & 1920      & -       \\
\texttt{height}                & 2855      & -       \\
\texttt{weight}                & 225       & -       \\
\bottomrule
\end{tabular}}
\end{table}

\begin{figure}[H]
  \centering
  \captionsetup{justification=centering}
  \includegraphics[width=1\columnwidth]{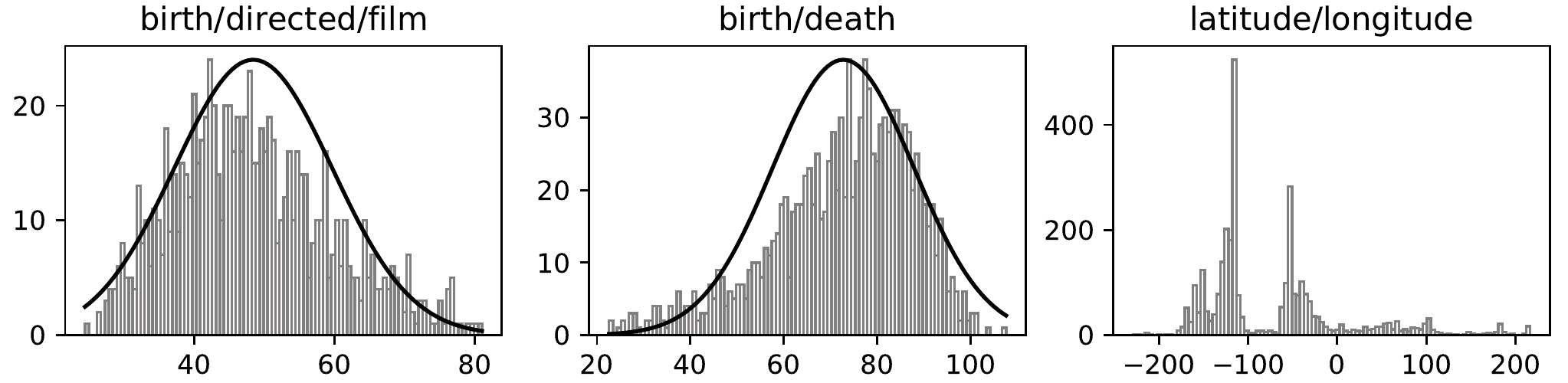}
	\caption{Histograms and fitted normal curves of node attribute differences computed along some relations}
  \label{fig:hist}
\end{figure}

We implement \textsc{MrAP}\footnote{Source code is available at https://github.com/bayrameda/MrAP} using the PyTorch-scatter package \cite{fey2019fast}, which provides an efficient computation of message passing on a sparse relational structure.
The damping factor of \textsc{MrAP} is set to $\xi = 0.5$, and the propagation stops upon reaching a convergence when the difference between two consequent iterations drops below $0.1\%$ of the range of attributes.
For the regression functions between a pair of attributes of the same type and the ones expressed in same numerical range and unit \eg, date attributes, the default value of parameter $\eta$ is $1$.
With this in mind, we plot the histograms of numerical attribute differences over some representative relationships in Figure \ref{fig:hist}. In the first two plots, we observe that the difference between \texttt{date\_of\_birth} of a person and \texttt{date\_release} of the film directed by that person easily fits a normal distribution as well as the difference between \texttt{date\_of\_birth} and \texttt{date\_of\_death} of a person. Here, the mean corresponds to the estimated value of the parameter $\tau$. The relation between these attributes conforms to the assumed linear regression model by our method.
On the other hand, \texttt{latitude} and \texttt{longitude} of a location do not accommodate such a correlation. Thus, \textsc{MrAP} can simply skip the message passing between such attributes. Given the number of attributes and relation types in each dataset, the total number of regression models actively used by \textsc{MrAP} is reported in Table \ref{tbl:stats}. Given also the number of multi-relational edges, it is possible to compute the number of message passing paths, which relates to the number of messages propagated across the graph in one iteration.
\begin{table}[h]
\centering
\caption{\label{tbl:stats} (Upper) Dataset statistics. (Lower) Characteristics of \textsc{MrAP} in these datasets.}
\resizebox{0.7\columnwidth}{!}{
\centering
\begin{tabular}{lrr}
\toprule
                     & FB15K-237 & YAGO15K \\
                     \midrule
Entities              & 10,054     & 15,077   \\
Edges      & 118,747    & 119,590  \\
Relation types        & 114       & 32      \\
Attribute types       & 11        & 7       \\
Attributes in tr.       & 9,261     & 9,405   \\
Attributes in dev.       &  2,315    & 2,351  \\
Attributes in test       & 2,315     & 2,351   \\\midrule
Message passing paths & 180,688    & 168,915  \\
Regression functions     & 310       & 261    \\
\bottomrule
\end{tabular}}
\end{table}

\xhdr{Baselines} We compare \textsc{MrAP} to baseline methods introduced in \cite{kotnis2018learning}: \textsc{Global} and \textsc{Local}. For each type of attribute, while \textsc{Global} replaces the missing values by the average of the known ones, \textsc{Local} replaces them by the average of the known ones in the neighboring nodes. We also compare to \textsc{Nap++} \cite{kotnis2018learning}. For each type of attribute, \textsc{Nap++} constructs a k-NN graph upon the learned node embedding solely for the propagation of that type of attribute. As opposed to these methods, \textsc{MrAP} leverages the correlations across all attribute types and the multi-relational structure of the KG to impute the missing values.

\xhdr{Experimental setup} Given KG datasets, we randomly split their node attributes into training, validation and test sets in a proportion of $80/10/10\%$. The validation set is used for the hyper-parameter tuning of \textsc{Nap++} framework and we measure the performance of all methods on the test set. Statistics for this configuration are summarized in Table \ref{tbl:stats}.
We run experiments on several setups with different sparsity of observed node attributes. For this purpose, we use randomly subsampled versions of the training set as observed attributes and we set the rest as missing. In this paper, we report the results for two different setups: in the former, we use all of the training set as observed attributes and in the latter, we target a higher regime of sparsity and we use half of the training set as observed attributes. Throughout the section, we refer to these setups as `100\%' and `50\%' respectively.

\xhdr{Analysis}
The performances of the methods on the two KG datasets are given in Table \ref{tbl: result_FB} and \ref{tbl: result_YAGO}. We see that the comparison of the methods across different setups (100\% and 50\%) is quite consistent. \textsc{MrAP}  achieves competitive results against the other methods, specifically on date type of attributes, it performs mostly the best in both of the two datasets.
We argue that this is achieved because \textsc{MrAP} profits the message passing between different types of attributes, unlike the other methods, which do not permit a direct information exchange between them.
This is found to be critical particularly among the date attributes: when the message passing between different types of attributes is deactivated in \textsc{MrAP}, the prediction
error for most of the date attributes raises.
We run additional experiments to justify other design choices of \textsc{MrAP}, and provide an ablation study in Table \ref{tbl: Ablation}. First, we refer to the case where the message passing between different types of attributes is deactivated as `w/o Cross' since this case blocks the information crossing from one attribute type to another. Second, we block the propagation of messages within a node, achieved by the inner loss term introduced in \eqref{eq:loss}, and we refer to this case as `w/o Inner'.
Note that the former case, `w/o Cross', already spans the latter, `w/o Inner', because the inner-node message passing is always realized between different types of attributes.
The experiments show that the cross-attribute and inner-node message passing enhances the prediction results almost always. We see that the inner-node message passing is significant in particular between the attributes \texttt{date\_of\_birth} and \texttt{date\_of\_death}, \texttt{area} and \texttt{population}, and then, \texttt{height} and \texttt{weight}. For instance, in the case `w/o Inner', the error for the attribute \texttt{date\_of\_death} raises more than $10\%$ as seen in Table \ref{tbl: Ablation}.
\begin{table}[H]
\centering
\caption{\label{tbl: Ablation}Ablation study for \textsc{MrAP}. MAE measured on the experimental setup `50\%'.}
\resizebox{\columnwidth}{!}{
\begin{tabular}{lp{3cm}rrr}
\toprule
    Dataset     &Attribute             & w/o Cross & w/o Inner &  \textsc{MrAP}\\
\midrule             
{\multirow{11}{*}{\parbox{1.2cm}{FB15K-237}}}
&\texttt{date\_of\_birth}       & 19.1      & 14.4  & \textbf{12.3} \\
&\texttt{date\_of\_death}       & 41.0      & 20.0  & \textbf{16.0} \\
&\texttt{film\_release}         & 11.5      & 6.4     &\textbf{6.4}   \\
&\texttt{organization\_founded} & 71.0      & \textbf{60.5}    & 60.9   \\
&\texttt{location\_founded}     & 148.7       & 106.1     & \textbf{105.9}   \\
&\texttt{latitude}              & 2.1      & 2.1  & \textbf{2.1} \\
&\texttt{longitude}             & 4.7      & 4.7  & \textbf{4.7} \\
&\texttt{area}                  & 1.8e6      & 1.8e6     & \textbf{5.7e5}  \\
&\texttt{population}            & 2.4e7      & 2.4e7     & \textbf{2.3e7}   \\
&\texttt{height}                & 0.089      & 0.089     & \textbf{0.087}    \\
&\texttt{weight}                & 16.6       & 16.6     & \textbf{13.2}    \\
\midrule
{\multirow{7}{*}{\parbox{1.2cm}{YAGO15K}}}
&\texttt{date\_of\_birth}       & 28.7      & 22.8  & \textbf{21.1} \\
&\texttt{date\_of\_death}       & 52.4      & 42.7  & \textbf{35.0} \\
&\texttt{date\_created}         & 86.8         & 65.9  & \textbf{65.8} \\
&\texttt{date\_destroyed}       & 43.3         & 30.4   & \textbf{28.1}  \\
&\texttt{date\_happened}        & 60.1         & 54.2   & \textbf{54.0}  \\
&\texttt{latitude}              & 3.7      & 3.7  & \textbf{3.7} \\
&\texttt{longitude}             & 7.4      & 7.4  & \textbf{7.4} \\
\bottomrule
\end{tabular}
}
\end{table}

\begin{table*}[h]
\centering
\caption{\label{tbl: result_FB}Performances on FB15K-237 with two different setup of observed node attribute sparsity}
\resizebox{1.9\columnwidth}{!}{
\begin{tabular}{lrrrrrrrrrrrr} \toprule
& \multicolumn{6}{c}{100 \%} & \multicolumn{6}{c}{50 \%} \\
\cmidrule[\heavyrulewidth](lr){2-7} \cmidrule[\heavyrulewidth](lr){8-13} 
      & \multicolumn{2}{c}{\textsc{Local/Global}} & \multicolumn{2}{c}{\textsc{Nap++}} & \multicolumn{2}{c}{\textsc{MrAP}}
      & \multicolumn{2}{c}{\textsc{Local/Global}} & \multicolumn{2}{c}{\textsc{Nap++}} & \multicolumn{2}{c}{\textsc{MrAP}}\\
      \cmidrule(lr){2-3} \cmidrule(lr){4-5} \cmidrule(lr){6-7} \cmidrule(lr){8-9} \cmidrule(lr){10-11} \cmidrule(lr){12-13}
Attribute       & \textbf{MAE}         & \textbf{RMSE}        & \textbf{MAE}         & \textbf{RMSE}        & \textbf{MAE}            & \textbf{RMSE}
& \textbf{MAE}         & \textbf{RMSE}        & \textbf{MAE}         & \textbf{RMSE}        & \textbf{MAE}            & \textbf{RMSE}\\
      \midrule
\texttt{date\_of\_birth}        & 20.6    & 54.2  & 22.1  & \textbf{34.3}  & \textbf{15.0}  & 38.6    & 24.0    & 69.4  & 27.2  & 40.0  & \textbf{12.3}  & \textbf{20.5}   \\
\texttt{date\_of\_death}        & 37.2    & 68.4  & 52.3  & 85.2  & \textbf{16.3}  & \textbf{32.2.2}    & 36.8    & 54.7  & 79.3  & 95.7  & \textbf{16.0}  & \textbf{25.2}     \\
\texttt{film\_release}         & 11.5    & 15.5  & 9.9   & 14.7  & \textbf{6.3}   & \textbf{8.6}     & 11.8    & 15.2  & 9.3   & 12.8  & \textbf{6.4}   & \textbf{9.0}     \\
\texttt{organization\_founded} & *73.3    & *121.0 & 59.3  & 98.0  & \textbf{58.3}  & \textbf{91.6}    & *72.3    & *121.4 & 65.0  & 114.6  & \textbf{60.9}  & \textbf{96.5} \\
\texttt{location\_founded}     & 138.0   & *259.8 & 149.9 & 277.0 & \textbf{98.8}  & \textbf{151.9}   & 111.7   & 176.4 & 165.4 & 291.7 & \textbf{105.9}  & \textbf{146.2} \\
\texttt{latitude}   & 3.3 & 10.3 & 11.8 & 18.9  & \textbf{1.5} & \textbf{3.5}     & 5.2 & 11.9 & 11.5 & 18.7  & \textbf{2.1} & \textbf{4.1}\\
\texttt{longitude}  & 6.2 & 16.3 & 54.7 & 71.8& \textbf{4.0} & \textbf{8.8}     & 22.4 & 38.4 & 51.7 & 66.9 & \textbf{4.7} & \textbf{9.3} \\
\texttt{area}       & *5.4e5 & *\textbf{5.4e5} & \textbf{4.4e5}   & 1.2e6  & 4.4e5 & 1.1e6 & *4.0e5 & *\textbf{4.1e5} & \textbf{3.2e5}   & 2.2e6  & 5.7e5 & 1.5e6 \\
\texttt{population} & *7.7e6  & *\textbf{1.8e7} & \textbf{7.5e6}   & 6.5e7  & 2.1e7 & 4.3e7 & *\textbf{5.0e6}  & *\textbf{1.8e7} & 7.5e6   & 6.4e7  & 2.3e7 & 4.2e7 \\
\texttt{height}     & *0.085      & *0.104     & \textbf{0.080}       & \textbf{0.102}      & 0.086     & 0.106     & *0.085      & *0.104     & \textbf{0.080}       & \textbf{0.102}      & 0.087     & 0.108 \\
\texttt{weight}     & *14.2     & *20.2    & 15.3      & 18.9     & \textbf{12.9}    & \textbf{18.3}    & *14.2     & *20.2    & 13.6      & \textbf{17.3}     & \textbf{13.2}    & 19.3  \\ 
\bottomrule
\end{tabular}
}
\end{table*}
\begin{table*}[h]
\centering
\caption{\label{tbl: result_YAGO} Performances on YAGO15K with two different setup of observed node attribute sparsity}
\resizebox{1.9\columnwidth}{!}{
\begin{tabular}{lrrrrrrrrrrrr} \toprule
& \multicolumn{6}{c}{100 \%} & \multicolumn{6}{c}{50 \%} \\
\cmidrule[\heavyrulewidth](lr){2-7} \cmidrule[\heavyrulewidth](lr){8-13} 
      & \multicolumn{2}{c}{\textsc{Local/Global}} & \multicolumn{2}{c}{\textsc{Nap++}} & \multicolumn{2}{c}{\textsc{MrAP}}
      & \multicolumn{2}{c}{\textsc{Local/Global}} & \multicolumn{2}{c}{\textsc{Nap++}} & \multicolumn{2}{c}{\textsc{MrAP}}\\
      \cmidrule(lr){2-3} \cmidrule(lr){4-5} \cmidrule(lr){6-7} \cmidrule(lr){8-9} \cmidrule(lr){10-11} \cmidrule(lr){12-13}
Attribute       & \textbf{MAE}         & \textbf{RMSE}        & \textbf{MAE}         & \textbf{RMSE}        & \textbf{MAE}            & \textbf{RMSE}
& \textbf{MAE}         & \textbf{RMSE}        & \textbf{MAE}         & \textbf{RMSE}        & \textbf{MAE}            & \textbf{RMSE}\\
      \midrule
\texttt{date\_of\_birth}        & 26.3    & 64.8  & 23.2  & 59.9  & \textbf{19.7}  & \textbf{31.5}    & 26.2 & 65.2 & 24.2  & \textbf{61.3}  & \textbf{21.1} & 61.9      \\
\texttt{date\_of\_death}        & *48.6    & *89.5  & 45.7  & 99.4  & \textbf{34.0}  & \textbf{84.2}    & *45.4  & *89.1 & 47.4 & 97.8  & \textbf{35.0} & \textbf{84.4}      \\
\texttt{date\_created}    & *95.5     & *155.8      & 83.5       & 152.3    & \textbf{70.4}  & \textbf{149.6}             & *96.0 & *155.8  & 82.6  & 152.6 & \textbf{65.8} & \textbf{135.3}\\
\texttt{date\_destroyed}  & 42.2       & \textbf{59.5}       & 38.2       & 75.5       & \textbf{34.6} & 62.0  & 41.8  & 59.3  & 33.9  & 68.3 & \textbf{28.1}    & \textbf{45.9}          \\
\texttt{date\_happened}   & *\textbf{52.1}       & *\textbf{67.3}       & 73.7       & 159.9      & 54.1  & 73.8  & *60.1  & *\textbf{72.7}  & 77.0 & 141.5      & \textbf{54.0} & 95.6          \\
\texttt{latitude}   & 3.4 & 9.0 & 8.7 & 13.8  & \textbf{2.8} & \textbf{7.9}     & 6.7  & 14.7  & 9.2 & 14.2  & \textbf{3.7} & \textbf{8.6}          \\
\texttt{longitude}  & 10.6 & 24.1 & 43.1 & 58.6 & \textbf{5.7} & \textbf{17.1}     & 20.5  & 34.6 & 45.2 & 60.9 & \textbf{7.4} & \textbf{18.0}          \\
\bottomrule
\end{tabular}
}
\end{table*}
In Table \ref{tbl: result_FB} and \ref{tbl: result_YAGO}, \textsc{Local/Global} reports the best performance obtained by either of the two baselines for each attribute and an asterisk (*) indicates that \textsc{Global} outperforms \textsc{Local}. We see that \textsc{Global} performs the best for some types of attributes, \eg, area and population. For the prediction of those, we argue that
the underlying relational structure may not be very informative, since the relation based methods, \ie, \textsc{Local}, \textsc{Nap++}, \textsc{MrAP}, perform poorly.
The attributes with least number of samples (see Table \ref{tbl: num_atts})
may also challenge the model parameter learning in \textsc{Nap++} and \textsc{MrAP} and affect their performance.
In addition, \textsc{Global} outperforms \textsc{Local} occasionally, \eg, \texttt{date\_organization\_founded} in FB15K-237 and \texttt{date\_created} in YAGO15K. Even if the relational structure underlying those attributes are informative, \textsc{Local} applies the neighborhood averaging regardless of the relation types. Here, \textsc{MrAP} improves the prediction by inducing relation and attribute specific regression models.
 
Besides a better overall performance, \textsc{MrAP} exhibits other advantages with respect to \textsc{Nap++}: while \textsc{MrAP} performs the estimation of its parameters and the imputation of the missing values in seconds, \textsc{Nap++} requires several hours, mostly due to the learning of node embeddings. The experiments are executed in a GTX Titan GPU.
\textsc{MrAP} is also more efficient in memory---it only has to learn three parameters per regression function---as compared to \textsc{Nap++}, which learns a latent representation (whose dimensionality is 100) per node.

\section{Conclusion}
\label{sec: Conclusion}
We address a relatively unexplored problem, node attribute completion, in knowledge graphs, and present \textsc{MrAP}, a multi-relational propagation algorithm to predict the missing node attributes. \textsc{MrAP} is framed in a message passing scheme, enabling the propagation of information across multiple types of attributes and over multiple types of relations. We show that \textsc{MrAP} very often outperforms several baselines in two datasets. Future work will focus on simultaneously learning the parameters of the regression functions while propagating the attributes in the knowledge graph.

\section{ACKNOWLEDGMENT}
We would like to thank Pierre Vandergheynst for supporting the project with his constructive comments, Bhushan Kotnis for sharing the code of the algorithm \textsc{NAP++} and Elif Vural for her helpful feedback on the paper.

\bibliographystyle{IEEEbib}
\bibliography{strings,refs}

\begin{thebibliography}{10}

\bibitem{west2014knowledge}
Robert West, Evgeniy Gabrilovich, Kevin Murphy, Shaohua Sun, Rahul Gupta, and
  Dekang Lin,
\newblock ``Knowledge base completion via search-based question answering,''
\newblock in {\em Proceedings of the 23rd international conference on World
  wide web}, 2014, pp. 515--526.

\bibitem{bordes2014question}
Antoine Bordes, Sumit Chopra, and Jason Weston,
\newblock ``Question answering with subgraph embeddings,''
\newblock in {\em Proceedings of the 2014 Conference on Empirical Methods in
  Natural Language Processing (EMNLP)}, 2014, pp. 615--620.

\bibitem{mohamed2020discovering}
Sameh~K Mohamed, V{\'\i}t Nov{\'a}{\v{c}}ek, and Aayah Nounu,
\newblock ``Discovering protein drug targets using knowledge graph
  embeddings,''
\newblock {\em Bioinformatics}, vol. 36, no. 2, pp. 603--610, 2020.

\bibitem{ioannidis2020few}
Vassilis~N Ioannidis, Da~Zheng, and George Karypis,
\newblock ``Few-shot link prediction via graph neural networks for covid-19
  drug-repurposing,''
\newblock {\em arXiv preprint arXiv:2007.10261}, 2020.

\bibitem{xu2020product}
Da~Xu, Chuanwei Ruan, Evren Korpeoglu, Sushant Kumar, and Kannan Achan,
\newblock ``Product knowledge graph embedding for e-commerce,''
\newblock in {\em Proceedings of the 13th International Conference on Web
  Search and Data Mining}, 2020, pp. 672--680.

\bibitem{li2020alime}
Feng-Lin Li, Hehong Chen, Guohai Xu, Tian Qiu, Feng Ji, Ji~Zhang, and Haiqing
  Chen,
\newblock ``Alime kg: Domain knowledge graph construction and application in
  e-commerce,''
\newblock {\em arXiv preprint arXiv:2009.11684}, 2020.

\bibitem{nickel2015review}
Maximilian Nickel, Kevin Murphy, Volker Tresp, and Evgeniy Gabrilovich,
\newblock ``A review of relational machine learning for knowledge graphs,''
\newblock {\em Proceedings of the IEEE}, vol. 104, no. 1, pp. 11--33, 2015.

\bibitem{wang2017knowledge}
Quan Wang, Zhendong Mao, Bin Wang, and Li~Guo,
\newblock ``Knowledge graph embedding: A survey of approaches and
  applications,''
\newblock {\em IEEE Transactions on Knowledge and Data Engineering}, vol. 29,
  no. 12, pp. 2724--2743, 2017.

\bibitem{scarselli2008graph}
Franco Scarselli, Marco Gori, Ah~Chung Tsoi, Markus Hagenbuchner, and Gabriele
  Monfardini,
\newblock ``The graph neural network model,''
\newblock {\em IEEE Transactions on Neural Networks}, vol. 20, no. 1, pp.
  61--80, 2008.

\bibitem{gilmer2017neural}
Justin Gilmer, Samuel~S Schoenholz, Patrick~F Riley, Oriol Vinyals, and
  George~E Dahl,
\newblock ``Neural message passing for quantum chemistry,''
\newblock {\em arXiv preprint arXiv:1704.01212}, 2017.

\bibitem{schlichtkrull2018modeling}
Michael Schlichtkrull, Thomas~N Kipf, Peter Bloem, Rianne Van Den~Berg, Ivan
  Titov, and Max Welling,
\newblock ``Modeling relational data with graph convolutional networks,''
\newblock in {\em European Semantic Web Conference}. Springer, 2018, pp.
  593--607.

\bibitem{garcia2017kblrn}
Alberto Garcia-Duran and Mathias Niepert,
\newblock ``Kblrn: End-to-end learning of knowledge base representations with
  latent, relational, and numerical features,''
\newblock in {\em Proceedings of the 34th Conference on Uncertainty in
  Artificial Intelligence}, 2018.

\bibitem{kotnis2018learning}
Bhushan Kotnis and Alberto Garc{\'\i}a-Dur{\'a}n,
\newblock ``Learning numerical attributes in knowledge bases,''
\newblock in {\em Automated Knowledge Base Construction (AKBC)}, 2018.

\bibitem{shang2019end}
Chao Shang, Yun Tang, Jing Huang, Jinbo Bi, Xiaodong He, and Bowen Zhou,
\newblock ``End-to-end structure-aware convolutional networks for knowledge
  base completion,''
\newblock in {\em Proceedings of the AAAI Conference on Artificial
  Intelligence}, 2019, vol.~33, pp. 3060--3067.

\bibitem{zhu2002learning}
Xiaojin Zhu and Zoubin Ghahramani,
\newblock ``Learning from labeled and unlabeled data with label propagation,''
\newblock {\em Tech. Rep., Technical Report CMU-CALD-02--107, Carnegie Mellon
  University}, 2002.

\bibitem{li2015gated}
Y.~Li, Daniel Tarlow, Marc Brockschmidt, and R.~Zemel,
\newblock ``Gated graph sequence neural networks,''
\newblock {\em CoRR}, vol. abs/1511.05493, 2016.

\bibitem{hamaguchi2017knowledge}
Takuo Hamaguchi, Hidekazu Oiwa, Masashi Shimbo, and Yuji Matsumoto,
\newblock ``Knowledge transfer for out-of-knowledge-base entities: a graph
  neural network approach,''
\newblock in {\em Proceedings of the 26th International Joint Conference on
  Artificial Intelligence}, 2017, pp. 1802--1808.

\bibitem{beck2018graph}
Daniel Beck, Gholamreza Haffari, and Trevor Cohn,
\newblock ``Graph-to-sequence learning using gated graph neural networks,''
\newblock in {\em Proceedings of the 56th Annual Meeting of the Association for
  Computational Linguistics (Volume 1: Long Papers)}, 2018, pp. 273--283.

\bibitem{brockschmidt2019gnn}
Marc Brockschmidt,
\newblock ``Gnn-film: Graph neural networks with feature-wise linear
  modulation,''
\newblock {\em arXiv preprint arXiv:1906.12192}, 2019.

\bibitem{teru2019inductive}
Komal~K Teru, Etienne Denis, and William~L Hamilton,
\newblock ``Inductive relation prediction by subgraph reasoning,''
\newblock {\em arXiv}, pp. arXiv--1911, 2019.

\bibitem{vashishth2019composition}
Shikhar Vashishth, Soumya Sanyal, Vikram Nitin, and Partha Talukdar,
\newblock ``Composition-based multi-relational graph convolutional networks,''
\newblock in {\em International Conference on Learning Representations}, 2020.

\bibitem{yu2020generalized}
Donghan Yu, Yiming Yang, Ruohong Zhang, and Yuexin Wu,
\newblock ``Generalized multi-relational graph convolution network,''
\newblock {\em arXiv preprint arXiv:2006.07331}, 2020.

\bibitem{neil2018interpretable}
Daniel Neil, Joss Briody, Alix Lacoste, Aaron Sim, Paidi Creed, and Amir
  Saffari,
\newblock ``Interpretable graph convolutional neural networks for inference on
  noisy knowledge graphs,''
\newblock {\em arXiv preprint arXiv:1812.00279}, 2018.

\bibitem{wang2019kgat}
Xiang Wang, Xiangnan He, Yixin Cao, Meng Liu, and Tat-Seng Chua,
\newblock ``Kgat: Knowledge graph attention network for recommendation,''
\newblock in {\em Proceedings of the 25th ACM SIGKDD International Conference
  on Knowledge Discovery \& Data Mining}, 2019, pp. 950--958.

\bibitem{busbridge2019relational}
Dan Busbridge, Dane Sherburn, Pietro Cavallo, and Nils~Y Hammerla,
\newblock ``Relational graph attention networks,''
\newblock {\em arXiv preprint arXiv:1904.05811}, 2019.

\bibitem{bayram2020mask}
Eda Bayram, Dorina Thanou, Elif Vural, and Pascal Frossard,
\newblock ``Mask combination of multi-layer graphs for global structure
  inference,''
\newblock {\em IEEE Transactions on Signal and Information Processing over
  Networks}, vol. 6, pp. 394--406, 2020.

\bibitem{hu2020heterogeneous}
Ziniu Hu, Yuxiao Dong, Kuansan Wang, and Yizhou Sun,
\newblock ``Heterogeneous graph transformer,''
\newblock in {\em Proceedings of The Web Conference 2020}, 2020, pp.
  2704--2710.

\bibitem{xu2018representation}
Keyulu Xu, Chengtao Li, Yonglong Tian, Tomohiro Sonobe, Ken-ichi Kawarabayashi,
  and Stefanie Jegelka,
\newblock ``Representation learning on graphs with jumping knowledge
  networks,''
\newblock in {\em International Conference on Machine Learning}, 2018, pp.
  5453--5462.

\bibitem{rencher2012}
A.~C. Rencher and W.~Christensen,
\newblock {\em Methods of Multivariate Analysis}, chapter~3, pp. 47--90,
\newblock John Wiley \& Sons, 2012.

\bibitem{toutanova2015observed}
Kristina Toutanova and Danqi Chen,
\newblock ``Observed versus latent features for knowledge base and text
  inference,''
\newblock in {\em Proceedings of the 3rd Workshop on Continuous Vector Space
  Models and their Compositionality}, 2015, pp. 57--66.

\bibitem{garcia2018learning}
Alberto Garcia-Duran, Sebastijan Duman{\v{c}}i{\'c}, and Mathias Niepert,
\newblock ``Learning sequence encoders for temporal knowledge graph
  completion,''
\newblock in {\em Proceedings of the 2018 Conference on Empirical Methods in
  Natural Language Processing}, 2018, pp. 4816--4821.

\bibitem{fey2019fast}
Matthias Fey and Jan~Eric Lenssen,
\newblock ``Fast graph representation learning with pytorch geometric,''
\newblock {\em arXiv preprint arXiv:1903.02428}, 2019.

\end{thebibliography}

\end{document}